# Sparse Auto-Encoders and Holism about Large Language Models

Jumbly Grindrod

University of Reading

0. Introduction

Does Large Language Model (LLM) technology suggest a meta-semantic picture i.e. a picture of how words and complex expressions come to have the meaning that they do? One modest approach explores the assumptions that seem to be built into how LLMs capture the meanings of linguistic expressions as a way of considering their plausibility (Grindrod, 2026a, 2026b). It has previously been argued that LLMs, in employing a form of distributional semantics, adopt a form of holism about meaning (Grindrod, 2023; Grindrod et al., forthcoming). However, recent work in mechanistic interpretability presents a challenge to these arguments. Specifically, the discovery of a vast array of interpretable latent features within the high dimensional spaces used by LLMs potentially challenges the holistic interpretation. In this paper, I will present the original reasons for thinking that LLMs embody a form of holism (section 1), before introducing recent work on features generated through *sparse auto-encoders*, and explaining how the discovery of such features suggests an alternative decompositional picture of meaning (section 2). I will then respond to this challenge by considering in greater detail the nature of such features (section 3). Finally, I will return to the holistic picture defended by Grindrod et al. and argue that the picture still stands provided that the features are countable (section 4).

1. LLM technology and holism about meaning

How is it that LLM technology could be related to philosophical questions about linguistic meaning? One approach to take is to claim that the astounding technological progress that LLMs have given rise to gives us good reason to think that the way that LLMs capture linguistic meaning must be accurate, at least broadly. This is so even if this requires us to reject assumptions about language and meaning that have previously been widely accepted. Of course, we need to keep this at the level of a broad generalization because there are many impressive LLMs today that will differ in various ways to one another. But if we can identify some of the significant technological developments that have helped better capture the meanings of linguistic expressions, and that are common across all or most LLMs today, this can give us real insight into the nature of linguistic meaning. This bold approach does have its advocates (Piantadosi, 2023), but of course it is controversial. A weaker approach to take is just to treat the way that LLMs capture meaning as indicative of a particular picture of linguistic meaning but hold off claiming that as a result the picture must be correct. Perhaps the success of LLM technology should increase our credence in the picture somewhat, or perhaps the picture is worthy of study even if we don't think it can be a correct theory of meaning – as it may be that it leads to change our reasoning as to why such a picture does not work. For the purposes of this paper, I am committed to this weaker approach. As such, in what follows I will discuss the *LLM picture of meaning*, which will refer to the assumptions about meaning that seem to be built into how LLMs work. As I've already noted, there are important variations across LLMs, but the discussion here will largely not focus on such differences. Those who prefer the strong approach can view this as directly informing meta-semantic debates. Those who prefer the weaker approach can view this merely as a study of how LLMs process meaning and what points of contact there are with central debates in philosophy of language.

It will do to start with a brief overview of how LLMs capture the meanings of words, and to do so, it is best to start with language models before they got large. Following some initial work from (Bengio, 2008) and others, the idea of using neural networks trained on a language prediction task really took hold with the introduction of Word2vec and similar models (Mikolov, Yih, et al., 2013; Mikolov, Chen, et al., 2013). Word2vec is a neural network with only a single hidden layer, with the number of neurons at that layer an adjustable parameter (around 300 is a standard setting). The input layer has a neuron for each word in the vocabulary, as does the output layer. Given some input text, the neurons in the input layer corresponding to each word in the text will individually fire, with the average of the various activation patterns used to generate output activations. Once converted into probabilities, these can then be used to generate predictions as to which word is missing, given the input text that surrounds the missing word.[1] During training, this task is done repeatedly for a large training corpus, with some loss function plus backpropagation used to adjust the model for each iteration in order to minimize error.

Each word can then be understood as represented via the weight connections between its neuron in the input layer and the weights in the hidden layer. Alternatively put, each word is represented by an embedding in a high dimensional space, where the elements of that embedding correspond to each weight attached to the input layer neuron. And much like the distributional models that largely preceded neural network models, the embedding space is ordered by *meaning proximity*: words that are more similar in meaning are closer together in the space. Within distributional semantics, this is known as the *distributional hypothesis*: that words with (dis)similar distributions across a suitably representative corpus will have (dis)similar meanings (Firth, 1957; Grindrod, 2023; Lenci, 2008). While neural network models are not explicitly distributional models in that their dimensions do not straightforwardly correspond to particular distributional properties, there is empirical evidence that Word2vec implicitly encode distributional patterns (Levy & Goldberg, 2014). Perhaps more importantly, there is in a sense only distributional information that is directly available to such models; after all, they trained (in their pre-training at least), only on raw, unannotated linguistic data.

---

[1] Here I have described the CBOW algorithm, which will predict which word is missing given the words that surround it. Word2vec can also be run with the skip-gram algorithm, which will predict the surround of a given word. Skip-gram is thought to be more effective for rarer words. I've focused on CBOW just because it more closely resembles the autoregressive models such as transformers (which predict a word given the words that have come prior).

The idea that the meaning of a word could be represented in a high-dimensional space, by representing meaning proximity relations, is quite a striking thesis. In philosophy, many are used to truth-conditional approaches that will represent the meaning of a word via its extension, where its extension may be a set of objects, or a function that takes in such sets and outputs something else, like another function or a truth value. The truth-conditional picture may be intensionalized in something like a possible world semantics, but still the core insight that we need to capture meaning via worldly relations remains the same. It is hard to find equivalent insights in the meaning spaces employed by language models such as Word2vec. For some, this leads us to a kind of grounding problem, where the representations of each word are not appropriately grounded in the actual world so as to completely capture word meanings (Coelho Mollo & Millière, 2023). Rather than capturing the meaning of each word via word-to-world relations, it seems instead that this approach captures the meaning of a word by its proximity to other words in the space. On that basis, (Grindrod et al., forthcoming) argue that such systems are holistic insofar as they meet Dresner's (2012, p. 611) definition:

> Relations of a certain kind (or kinds) that obtain among expressions of a given natural language (all of these expressions, or many of them) are constitutive for linguistic expressions to mean what they do.

What exactly the relations are that hold between words in a Word2vec space is not easily stated. The dimensions of the space are not interpretable and will stand for some very higher order properties of the distribution of each term with respect to one another. Grindrod et al. then defend this kind of holism from famous objections by Fodor and Lepore (Fodor & Lepore, 1992, 1999) regarding instability. The crux of Fodor and Lepore's objections is that they argue there is no reputable notion of meaning similarity that holistic theories could make use of. A distinctive feature of semantic space approaches though is that they allow for well-defined notions of similarity as proximity in the semantic space (whether that is cosine similarity, Euclidean similarity, city block similarity, or something else). But Fodor and Lepore argue that the problem recurs for such views in the question of how exactly the space is determined. Two words may be near in one space and distant in another, corresponding to two different measures of similarity. So how do we decide which space we use? A related question is whether we could have any similarity measure across spaces as well as within a space. A measure of distance within a space obviously gets us not closer to answer this question.

Grindrod et al. argue that the appropriate response to this objection requires seeing how meaning is treated by practitioners that make use of such semantic spaces. A very common practice is not to consider meaning as something that is defined within a semantic space but as something that all semantic spaces represent to a greater or lesser extent. Two semantic spaces are treated as capturing the meaning properties of a given word in the same way to the extent that the word has the same set of nearest neighbours within each space. This gives rise to what Grindrod et al. call the *differential view of meaning*. On this picture, it is not the word's location in a particular space that is representing the word's meaning, but the proximity relations to other words in the space. These high-dimensional spaces are really being used to capture a network of words, where the strength of relations in the networks is represented in the space by proximity. In this way, two distinct spaces (with distinct dimensions and even of different dimensionalities) could represent the same meaning relations that hold between the same set of words in the space.

Models like Word2vec are today dwarfed by LLMs, the vast majority of which employ some form of the transformer architecture (Vaswani et al., 2017). The transformer architecture provides a way of not merely representing each word type via an embedding, but each instance of every word within a given text via its own embedding. While models such as recurrent neural networks could achieve something similar, the transformer is able to do so without creating a training bottleneck by employing sequential processing (i.e. where each datapoint must be processed one after another in order). Because transformers avoid sequential processing, training much larger transformer models is more viable. So, with transformers came much larger models capable of a high degree of sensitivity to the textual context of every given word, and this led to huge leaps in progress that we saw from 2017 onwards, coupled with the later introductions of chatbots such as Chat-GPT, Gemini, and Claude.

For our purposes though, and as Grindrod et al. (forthcoming) argue, the points made above about how neural network models appear to embody a form of holism apply equally to transformer models. Transformer models still represent words in a high dimensional space which conforms to meaning proximity.[2] The key difference – beyond just the size of the models used to generate such embeddings – is that transformers use token embeddings as a starting point to produce *contextualised embeddings* i.e. embeddings that represent a word as used at a particular point in a wider text. This is achieved by running the initial token embeddings through many layers, each consisting in two sub-layers of self-attention heads and multi-layer perceptions (MLPs). The self-attention heads create three lower-dimensionality projections of the embedding it receives as input. The three projections are known as the *query*, *key,* and *value*. The query embedding for a given word is measured for similarity against all of the key embeddings for the surrounding words. Words whose keys received a high similarity score will receive a high -self-attention score. The self-attention scores are then multiplied against the value embeddings in a weighted sum in order to generate a new embedding. The resulting embedding is then added back to the residual stream.[3] The MLP is used to introduce non-linear computations that are not possible in the self-attention layers. There, the residual stream is projected into a higher-dimensional space then a non-linear function such as GeLU is applied, before it is then projected back down to the original dimensionality and added back to the residual stream. What exactly each sub-layer does is still an open question that is being explored extensively, but as a generalisation it is fair to say that many view the self-attention layers as tasked with passing information between the various token representations, while the MLPs are charged with introducing new information in order to modify how the various tokens are represented.[4]

---

[2] To some extent this is complicated by the fact that through the transformer procedure, a much wider array of information is introduced, including contextual information about the rest of the text but also worldly information.
[3] The residual stream is the continuous representation of the word in question, for which each layer of the transformer outputs some modification of.

The outcome of this procedure is still a set of embeddings that are used to represent the input text, so that next-word predictions can be assigned to the vocabulary. And just as with smaller models like Word2vec, a natural way to understand these embeddings is as sitting in a kind of meaning similarity network with regards to one another. The key step that the transformer architecture introduces is the chance to refine these representations so that they better capture the context of the wider text. But this step doesn't detract from the earlier reasoning towards a kind of holism.

2. Sparse auto-encoder features

So far, so holistic. But recent work in mechanistic interpretability seems to challenge this picture. To see why, we need to turn to some exciting recent work that has sought to better understand how transformer language models work.[5]

Mechanistic interpretability is an area of AI research that attempts to produce human-understandable interpretations of how LLMs produce the outputs that they do. Olah et al. (2020) proposed that the best way for mechanistic interpretability to proceed is through the identification of *features* and *circuits*.[6] Features can be understood for our purposes as neural network activation patterns that can be interpreted. For example, if a neural network fires in a particular always in response to images of clowns, then we can say that the neural activates its clown feature when it is fed those images. As Olah et al. detail, in many cases we would hope that the behaviour of easily delineable parts of the network correspond to feature activations e.g. a specific neuron fires when the model is given clown images. But instead, in LLMs we tend to find *polysemanticity*: individual neurons in LLMs can very rarely be straightforwardly interpreted as sensitive to some concept or notion. Instead, features more plausibly exist in LLMs as complex patterns of many neurons firing in concert. The reason for this is that the model is trying to encode more features than it has dimensions for (even in the higher-dimensional space found in the MLP). So, a given individual neuron's activity may bear a complex relationship to many different seemingly unrelated features. This is sometimes put by saying that features are in *superposition* (Elhage et al., 2022; Olah et al., 2020). In geometric terms, rather than treating a given dimension in our high-dimensional space as a feature, features exist as non-orthogonal directions in the space. Having features in superposition obviously makes interpretability more challenging, because we cannot just read off the features in a model by inspecting the behaviour of an individual neuron on each activation. It also introduces *interference*, where the activation of one feature will also activate other features to some lesser extent. This is the price that the model pays for being able to encode many more features than it has dimensions for.

---

[4] Against this generalisation: there is evidence that self-attention heads can store factual information in their parameters. For instance, (Geva et al., 2023) found that certain self-attention heads store subject-attribute relationships.
[5] For philosophical works that cover some of the same technical detail, see: (Beckmann & Queloz, forthcoming; Chalmers, n.d.).
[6] For the purposes of this paper, I will largely focus on features.

Recent work has sought to overcome this issue by producing models that can be used to interpret the activation patterns of an LLM. A common approach is to use *sparse auto-encoders* (Bricken et al., 2023). A sparse auto-encoder (SAE) is a neural network that is tasked with reproducing the activation patterns of a given layer an LLM. The SAE is a very wide and shallow neural network. It is shallow in that it has only a single hidden layer. But that hidden layer may be tens of thousands or even millions of neurons wide. Given the task of replicating its input as output, the network could just learn a copy function: copy the input into some part of the middle layer and copy it again as output. To prevent this, the model is encouraged to only use very sparse activations i.e. only a few of the middle layer neurons can fire at any given time. It is encouraged in this direction using L1 regularisation: the sum of the values of all middle layer neurons firing (or some multiplication of that figure) is added to the loss value at training. This means that more neurons firing will lead to worse performance in its training task. With the model discouraged from merely copying, the SAE is effectively encouraged to use the large number of neurons in its middle layer to capture the various possible patterns that arise in the LLM's activations. The hope would then be that the individual neurons in the SAE's middle layer are interpretable. And this is broadly what was found. For instance (Bricken et al., 2023) provide an interface where the thousands of features that their SAE identified can be studied,[7] with subsequent work producing even more impressive results at greater scale (Ameisen et al., 2025; Cunningham et al., 2023; Lindsey et al., 2025; Templeton et al., 2024).[8] These features can also be used to manipulate the behaviour of the LLM. The weights between the feature neuron in the middle layer of the SAE and the output layer capture how the feature is represented as a particular direction in the LLM's residual stream space, so the residual stream in the LLM can be "clamped" to that feature by just pushing it in that direction. In a famous example, Templeton et al. (2024) identified a *Golden Gate Bridge* feature in Claude and then steered Claude in that direction to the point that it would only talk about the Golden Gate Bridge and eve self-identified as the Golden Gate Bridge.[9]

---

[7] https://transformer-circuits.pub/2023/monosemantic-features/vis/a1.html

[8] Alongside SAEs, there are also *transcoders* (Dunefsky et al., 2024). Whereas SAEs attempt to reproduce their input as output, a transcoder will take the activation pattern that serves as input for a given sub-layer, and is tasked with reproducing as output the activation pattern that the sub-layer outputs. It is initially tempting to think that transcoders and SAEs are recreating different parts of the model: SAEs features will capture the representations of an LLM while transcoder features will capture the computation that the LLM performs. But in fact, there is significant overlap that one usually finds between the two. After all, both SAEs and transcoders face the initial task of encoding their input as a set of features. They then use those features differently (corresponding to differences between the hidden feature layer and the output layer). To the extent that the two do identify different sets of features, it is that transcoders will place greater focus on aspects of the activation pattern that are causally relevant in that particular sub-layer of the LLM. To use a toy example, if a sub-layer is particular focused on morphological properties, then the features produced by a transcoder are much more likely to be morphological. An SAE, on the other hand, would still be tasked with capturing all aspects of the input activation pattern, regardless of what the sub-layer focuses on at that level.

[9] Such steering is a somewhat artificial exercise, where in order to generate noticeable results, the feature has to be activated to a level that it would not usually reach. And steering will also often break the model, leading it to just

Subsequent work has sought to trace the causal effects of such features, ultimately identifying *circuits* that can go some way to providing explanations as to a model's behaviour across a range of outputs. Lindsey et al. (2025) and Ameisen et al. (2025) show that with a variant of an SAE known as a cross-layer transcoder, *attribution graphs* can be produced that go some way to revealing something like a thought process that a given LLM undergoes prior to producing the output that it does. For instance, in predicting the next word of a prompt like "The capital of the state that contains Dallas is…", an attribution graph can be used to reveal that an LLM will activate a capitals feature and a states feature, then a Texas feature, and this will then positively influence the prediction of "Austin" as the output.[10] Such findings are certainly exciting, but how does the discovery of such features challenge the holistic picture described in the previous section?

The discovery of these implicit features is, as many have noted, continuous with earlier findings. Famously, (Mikolov, Yih, et al., 2013) found that there seem to be latent features in the embedding space for their Word2vec models that can be identified through a kind of vector arithmetic. Using "[word]$_v$" to refer to the vector for "[word]", they found that king$_v$ – man$_v$ + woman$_v$ ≈ queen$_v$. This regularity persisted for other terms that are paired along gender lines e.g. boy/girl, uncle/aunt etc. This showed that there is a kind of gender direction in those spaces. But whereas these earlier findings found individual examples of latent features, the approach afforded by SAEs drives this to an extreme, where potentially millions of features are identified. This opens up the possibility of a picture where the meanings of words in the space can be understood as a composition of the basic lexicon of features. Perhaps word meanings are not captured via their geometric relationship with one another, but are instead captured as a combination of features. Perhaps the meaning of "king" is to be understood via features like *masculine, royal, leader, head-of-family* etc. rather than by its proximity to words like "queen", "prince", etc. This would serve to make sense not only of the meanings of token embeddings residing in the particular space that they do, but also the transformations that take place to produce contextualised embeddings. That is, we could now understand the advance that the transformer architecture brought as one where features could be added to or removed from contextualised embeddings. For instance, perhaps in the sentence "Edith got a toy lion for her birthday", the embedding for "lion" has features like *ferocious, dangerous,* and *large* removed and features like *cuddly* and *soft* etc. added in.

The idea of understanding word meanings via a basic lexicon of atomic meanings is reminiscent of decompositional views such as (Katz & Fodor, 1963). Such a position need not entail holism. It is tempting to think that the position cannot be holistic, for it captures the meanings of words not via word-to-word relations but via relations to some other set of entities i.e. the atomic meanings. But if it were the case that all words bare some relation to all basic meanings, we

---

produce nonsense or output the same word again and again. But this doesn't detract from the fact that in cases where steering works, it suggests that such features are playing a causal role in the production of the model's outputs.

[10] A demonstration of this can be found at: https://tinyurl.com/ydjmx6md

would arguably still have a form of holism at play, but where the inter-word relations in question make appeal to these basic meanings. It is for this reason that such a view is consistent with holism but doesn't entail it. Still, the way that most have imagined the view, including the likes of Katz and Fodor, is in a non-holistic sense where some words will bear no relation to some basic meanings (e.g. perhaps "bachelor bears some relation to the meaning *man*, but "breakfast" will not). Let's label the view that word meanings in LLMs are to be understood as composed out of features the *feature composition view*. This view will be the topic of the next section.

In the next section I want to consider whether the feature composition view could be correct. To do so, we should first address an argument that suggests it *must be*, which runs as follows. Provided that SAEs do succeed in reproducing the activation patterns of an LLM, the features that they identify as active for each activation pattern must be the components that the activation pattern is made up of. For the way that the activation patterns are replicated is *by combining the activation of many features all at once*. So for a given activation pattern that is replicated by features f1-fn, how could we deny that the activation pattern has the feature activations as its parts?

The problem with this argument is that it conflates the network-based properties of the activation patterns with the semantic properties that those activation patterns are supposed to possess. To see this, consider an analogy. Suppose that I get every English word printed on an a4 piece of paper, and I cut each piece into 10 evenly-spaced columns. Now it may be that many of my columns will be duplicates, that by combining the same column with various other columns I can reproduce many different words. Does this mean that I have identified some property common to all of those words? Yes, but this property is nowhere near being a semantic one. It looks to be a property specific to printing off words in a particular way, having a particular format and font, cutting up the resulting paper in a particular way etc. This process would reveal nothing to us about the semantics of the words. For the same reason, the mere fact that an activation pattern can be recreated using a set of SAE features alone doesn't guarantee that we have thereby identified some basic semantic parts. Of course, the fact that the features themselves appear to be interpretable *is* suggestive of the idea that we have identified some basic meaning constituents, and this is what the feature composition view has in its favour. In the next section, I will consider the plausibility of the feature composition view in greater detail.

3. The feature composition view

Given that the feature composition view closely resembles previous decompositional approaches, it is worth first considering previous philosophical objections against decompositional views. A relevant figure here is Jerry Fodor, who as mentioned defended a kind of decompositional view in his earlier work alongside Jerrold Katz. But he later changed his mind, and for much of his career argued that "definitional" views of meaning (where meaning is understood as analysable into basic meaning parts) cannot work. He presented a number of reasons for thinking this (many of which can be found in (Fodor, 1998)).

First, Quine (1951) famously argued that there is no non-circular account of the analytic/synthetic distinction and so any such distinction should be abandoned. This obviously speaks against the possibility of the definitional claims that would capture any decompositional approach. Second, the general project of producing definitions just looks like it hasn't produced many successes beyond "bachelor" and "vixen" (Fodor, 1998, p. 46). Third, empirical evidence in psycholinguistics and developmental psychology are not suggestive of concepts or word meanings having basic meanings as constituent parts (e.g. there is little evidence of children first gaining understanding of basic meanings and then composing them into complex meanings). Fourth, for certain meanings we face what Fodor calls the "residuum problem" (Fodor, 1998, p. 109). If complex meanings are to be understood as consisting in some set of meaning parts $M_1$, $M_2$,…$M_n$, then there should be a further possible meaning consisting in the same parts save one i.e. $M_1$, $M_2$, …$M_{n-1}$. But if we consider the meaning of "red" as an example, it is plausible that a component part of the meaning would be *colour*, for there is a clear entailment relation between whether something is red and whether something is coloured. So the meaning of "red" would be *colour* plus whatever else differentiates red from other colours – call this the *red residuum*. The problem is if we try to consider a meaning that only consists in the red residuum. How could this fail to pick out something that is not also a colour? But then it seems like the colour constituent is really redundant, that the red residuum is sufficient to pick out red alone. This then undercuts the initial motivation for identifying constituents in the first place.

Do these problems apply to the feature composition view? Taking the LLM picture of language seriously means that some of these objections can be addressed directly. Quine argued for a negative existential: that there is no non-circular distinction to be drawn between the analytic and the synthetic. Negative existentials are hard to prove though, and so one possible response is to claim that empirical inquiry will provide the basis for the distinction. Chomsky (2000, pp. 63–64) has previously taken this line: "The question of the existence of analytic truths and semantic connections more generally is an empirical one, to be settled by inquiry that goes well beyond the range of evidence ordinarily brought to bear in the literature on these topics".[11] And although Chomsky himself has made clear his distaste for anything like the LLM picture of language (Chomsky et al., 2023), the defender of the current proposal could argue that the discovery of features through the development of LLMs provides just such a basis.

Regarding the failure of the definitional project, the feature composition view may provide a remedy to such failures. Perhaps part of the reason why definitions have proven elusive is that previous forms of decompositional view have been oversimplified. Usually, a mereological picture is assumed where for any given meaning complex and any given candidate part (i.e. some basic meaning), a polar question can be asked as to whether the basis meaning is or is not part of the complex. The feature composition would instead suggest that if any feature plays a part in the construction of some meaning, it is by degree, according to the extent to which the embedding in question is displaced in the direction of that feature. Perhaps this more complex picture opens up the possibility of better capturing the definitional make-up of any given word. Furthermore, the very construction of LLMs themselves brings a radically different way of investigating meaning. Rather than relying on an armchair approach or even a survey-based approach, the introduction of LLMs provides us with the opportunity to take huge amounts of actual language use and construct meaning representations – embeddings and features – through unsupervised means. This would also go some way to giving an answer to the third worry from empirical considerations. While it may be that psycholinguistic or ontogenetic considerations in favour of the view are light on the ground, the empirical basis through which features are discovered is itself its own kind of evidence. Of course, eventually the commitment to such features would have to be reconciled with their lack of empirical support in other areas, but this should be treated as an open question rather than a knock-down objection.

---

[11] For more on Chomsky's views on the analytic/synthetic distinction, see: (Rey, 2023).

The residuum problem is interesting to consider in the context of the feature composition view. Fodor phrased the residuum problem in terms of impossible meanings: "if you take 'COLOUR' out of the definition of 'red', what you're left with *doesn't seem to be a possible meaning*; the residuum of 'red → coloured' is apparently a surd" (Fodor, 1998, p. 109, emphasis Fodor's). The problem compounds for the feature composition picture, for now we must consider not only the possibility that a feature might be altogether absent, but also that a feature might be activated merely to a lesser degree. To use our toy example, say the "king" embedding consists of activations of features *masculine, royal, leader, head-of-family*, all to some appropriate degree. What would it even mean to halve the activation strength of the *royal* feature, for instance? The idea of representing categorical notions on a scale also leads to a kind of impossible meaning. I leave the residuum problem here without endorsing any particular response to it.

While it is interesting to consider the kinds of problems that Fodor raises against decompositional views, the problems with the feature composition view arguably start to show up once we consider in greater detail how it might work. One practical problem is the earlier mentioned issue of interference. Recall that due to superposition, the activation of a given feature will activate other features to a lesser extent. Perhaps it is not possible to activate *royal* without activating a host of other unrelated features as well. But then we require some way of drawing a distinction between those features that are playing a significant role in the embedding representation versus those that are really just noise. Further investigation may yield such a distinction. It may be the case that feature activations at some low enough level are treated as noise by the model. Or it may be that the model has developed some more sophisticated ability to identify noise downstream of the assignment of features. Perhaps the most sensible way to discharge the worry is to treat it as a limitation of current models. As Elhage et al. (2022) argue, the use of superposition by such models is really an effort at simulating the behaviour of a larger model; for any model that exhibits polysemanticity, it is possible to conceive of a larger monosemantic model that computers the same function without exploiting superposition. For this larger model, there will be no interference.

Actual inspection of the kinds of features that SAEs tend to generate should lead us to question whether they could serve in this kind of decompositional role. One initial reason for thinking this is that, as Chalmers (n.d., pp. 18–19) has noted, many features do not correspond to meaningful units in the philosopher's sense, with many picking out syntactic, sentential, or morphological properties, while others pick out properties of the LLM (e.g. (Templeton et al., 2024) identified a sycophancy feature that fires when the model is being sycophantic). This is part-and-parcel though of the fact that LLMs are tasked with processing all aspects of a text through the same set of representations. Part of the LLM picture of meaning involves accepting that semantic properties are not captured through their own dedicated objects, but must be disentangled from other linguistic and worldly properties, whether that is in token embeddings, contextualised embeddings, or in a feature list.

We can get an idea of which features fire for a given text using the interactive visualization from (Bricken et al., 2023). This allows you to see which features fire for every token of a few different text samples. One cherry-picked example might an idea of the kinds of features one can expect to see. In the third sentence of Moby Dick, the sentence starts "Whenever I find myself growing grim in the mouth…". For the token "growing", there are only four features that contribute to the total activation at a level detectable to 2 decimal places:[12]

| Index | % of Max | Human Label | Autointerpreted Label |
| --- | --- | --- | --- |
| #347 | 45.44 | "grow"/ "grown" | Looking at the top activating examples, this neuron seems to fire on the word "grow" or inflections of grow (grew, growing, grows). It seems to fire more strongly when "grow" is used literally to refer to something increasing in size, height, quantity, etc. As the activations get weaker, the neuron still often activates on forms of "grow", but also begins to fire on some related words like "rise" and "expand". The positive logits also reflect this, with words like "louder", "rapidly", and "exponentially". The negative logits don't tell us much. So in summary, this neuron fires mainly on the word "grow" and its inflections, especially when used literally to refer to something increasing in size or quantity.</thinking> |
| #3139 | 12.43 | Word endings with "-ing" | The neuron attends to verbs ending in "-ing", with stronger activations for common verbs. |
| #3962 | 15.02 | Words predicting me/myself/I ? | This neuron seems to fire for personal first person pronouns, actions, and adverbs relating to the passage of time. In <quantile_top>, the neuron fires for words like "wanted", "kickstart", "my", and "period". These are all first person pronouns and actions. In <quantile_7>, the neuron fires for "roughly", "halfway", "through", "on", and "first". These are all words relating to the passage of time. In <quantile_5>, the neuron fires for "without", "any", and "intimidating". These are more generic adverbs and pronouns. In the lower quantiles like <quantile_1> and <quantile_0>, the neuron fires more randomly |

---

[12] As is common with SAE features, the features here are given an autointerpreted label, where an LLM is asked to look at the behaviour prior to and after the activation of the feature as a way of interpreting what the feature represents. Two key sets of information contribute to this. One is the top token sequences that trigger various levels of activation. The other is the tokens that the activation of this feature would make more or less likely to predict (the "positive logits" and "negative logits"). Less commonly, (Bricken et al., 2023) also provide shorter human labels.

| | | | on words like "spent", "when", and "end". The <positive_logits> reinforce this pattern, with words like "myself", "my", "me", and adverbs like "awhile" and "lately". So in summary, this neuron seems to fire on first person pronouns, actions, and adverbs relating to the passage of time.</thinking> |
|---|---|---|---|
| #3663 | 8.29 | Physical/ human-body-related verbs | The neuron fires on words describing physical actions and states involving force or effort. |

This case was selected because of the relatively few features that activated, making it a fairly clean example to inspect. Selecting such a clean example, I take it, is friendly to the feature composition view, as it gives us the best chance of seeing how the meaning of a word could decompose. It is important to keep in mind that many cases are not like this; usually more features are firing including some that are hard to interpret as part of the meaning. Turning to this case, we can see a way in which the four features combine to capture what "growing" is doing in the text. The top two features capture the particular word form at play here as the "-ing" modified form of the lemma "grow". The human label for the third feature doesn't capture the subtlety of what the feature fires for according to the autointerpreted label. The feature fires for verbs related to the passage of time and applied to the first-person, all of which are true for "growing" in this text. The fourth feature doesn't obviously apply according to the autointerpreted label, as growing grim about the mouth is arguably not a physical action or state of exertion. However, an inspection of the top activations for that feature suggests that the autointerpreted label may be inaccurate to an extent. The feature appears to fire for verbs that take body parts as one of its arguments e.g. "his knees flexed", "eyes fluttered shut", "her nostrils flared". The four features then do seem to capture facts about the role that "growing" plays in the text. But in another sense, this looks a long way off capturing the kind of decomposition that such compositional views are typically thought to provide. Here it seems that the first feature is doing much of the heavy lifting in capturing the core meaning of the token. This also echoes the residuum problem as it seems that the parts that we identify end up leaving little semantic room for any other parts.

Of course, this is only one example, and Bricken et al.'s paper produced features only for a single layer transformer. We can also consider the nature of the features by taking a random sample of their descriptions to see what they typically fire for. The following is a random sample of 5 features across 5 layers for the LLM gemma-2-2b using the SAE GEMMASCOPE_RES_16K:

| layer | feature | description |
|---|---|---|
| 0 | 560 | references to salt in various contexts |
| 0 | 14856 | statements and references in legal or formal contexts |
| 0 | 6674 | references to roofs or roofing materials |
| 0 | 4083 | terms related to bonuses and rewards |
| 0 | 10851 | references to air conditioning systems |
| 5 | 7599 | occurrences of the phrase "working on" along with related terms and phrases |
| 5 | 312 | terms related to sports injuries and their consequences |
| 5 | 15478 | terms related to reciprocal relationships or comparatives |

| 5  | 10420 | occurrences of the "namespace" keyword in code |
| 5  | 2807  | terms related to removal and purification processes in scientific contexts |
| 12 | 7745  | references to familial relationships and the emotions tied to them |
| 12 | 13070 | structured comments and annotations in code documentation |
| 12 | 86    | conditional statements related to functions and parameters in programming |
| 12 | 12831 | commands and scripting elements related to programming or system operations |
| 12 | 5181  | references to health-related issues and diseases |
| 20 | 2836  | numbers and their relationships |
| 20 | 4560  | expressions of strong emotions and pleas for help or boundaries |
| 20 | 14297 | descriptions of visual attention or focus |
| 20 | 16339 | legal case information and citations. |
| 20 | 13030 | negative values associated with companies or medical interventions |
| 25 | 13517 | references to mystical or spiritual experiences and their implications for personal transformation |
| 25 | 7874  | technical terms and descriptions related to video compression standards |
| 25 | 7827  | references to specific tech products and their features |
| 25 | 8341  | information related to aviation regulations and pilot credentials |
| 25 | 6523  | references to military campaigns and territorial control |

Something that should be clear about this list is that it certainly doesn't seem like any layers are exclusively tracking semantic categories that could be basic building blocks of more complex meanings. Rather, the features appear to be highly specific. This is partly a symptom of how they are produced. Born as they are out of mechanistic interpretability, there was initially a rush towards producing larger SAEs with more features, thus providing a finer grain of analysis of the LLM in question. For instance, where (Bricken et al., 2023) focused on an SAE with only 4,096 features, the follow-up paper by (Templeton et al., 2024) focused on SAEs of various sizes up to 34 million features. This is obviously because mechanistic interpretability is about providing the a detailed by comprehendible description of what an LLM is doing on any given inference, and so one way to get a more detailed and accurate picture is to encourage the SAE to make finer-grained distinctions between the various kinds of activation patterns it is tasked with replicating. But the more we increase the number of features being appealed to, the further away we move from the decompositional idea of identifying the meanings that word meanings consist in. Certainly, by the time we have reached tens of millions of features, the number of features comfortably outnumbers the number of words or tokens in the lexicon, and so the SAE faces no pressure to identify what is fundamental in such systems.

The SAE literature has increasingly recognized that the rush towards more features has led to related problems. One issue concerns *feature duplication*. To capture the behaviour of an entire LLM, separates SAEs need to be produced for each sub-layer of the LLM (three for each layer: one at the self-attention sub-layer, one at the MLP sub-layer, and one at the residual stream). For instance, the GEMMASCOPE SAE set appealed to above consists in 78 SAEs, all trained separately independently of one another. Each individual SAE will likely identify a large number of the same features. This problem can be overcome with a more complex architecture. (Ameisen et al., 2025; Lindsey et al., 2025) introduced *cross-layer transcoders* as a sophisticated variant of SAEs and transcoders. These will use the same set of features to reconstruct all levels of an LLM, with an activation pattern at a given layer reconstructed by summing the contributions of all feature activations at all layers prior to and including the current layer.

A separate issue concerns *feature absorption* (Bussmann et al., 2025). If we have one more general feature and one more specific feature, where latter applies in all cases where the former applies, the pressure towards sparsity that SAEs face sometimes lead them to an odd outcome. Take the case of *women's names* and *instances of the name "Lily"*. A sparser outcome for the SAE would be one feature firing on a given occasion rather than two, so the SAE will be pushed towards not learning the feature *women's names* but instead learning the feature *women's names that aren't "Lily"*. This is recognized as a problem for interpretability, but it also speaks against the idea that suitably generalised meaning basics can be identified by pushing a neural network towards sparsity. Sparsity is an imperfect proxy for meaning that leads the SAEs to not cut the meaning space by its joints.

A third issue concerns what is known as *feature composition* (Leask et al., 2025). As larger SAEs with more features are produced, sparsity again encourages them to learn features that are specific in unintuitive ways. Suppose that a smaller SAE might learn the five following features: *circle, square, red, blue, green*. The problem is again that if a text that refers to, say a blue square is processed, two of these features will fire at once. A sparse outcome is to have more complex features so that only one needs to fire at a time. So a larger SAE may well the following six features instead: *red circle, blue circle, green circle, red square, blue square, green square*. More features are required, but a larger SAE can afford that cost to achieve greater sparsity. Importantly for our purposes though, it seems that with a larger SAE we seem to be moving further away from what seem like the basic constituents of meaning.

There may well be engineering solutions to these problems that mitigate the negative effects of using sparsity as a proxy for interpretability. Bussman et al. (2025) introduced *Matryoshka SAEs* for this reason. Matryoshka SAEs, like the Russian dolls they are named after, encourage a kind of hierarchy of generality among their features. The features are split into, for example, 5 groups, where the groups are subject to a differing number of loss functions, corresponding to how many of the feature groups are used in to reconstruct the input activation. The result is that the feature groups that are subject to more loss functions are pressured towards generality, and this has been shown to counteract both feature absorption and feature composition. But whether these kinds of solutions can get us towards the kind of meaning basics one would expect to find according to the feature composition view remains to be seen. Here is a sample of 25 Matryoshka features across 5 different feature groups, from 12-RES-MATRYOSHKA-DC SAEs of Gemma-2-2b:[13]

---

[13] There are accessible via neuronpedia.org.

| Loss functions | index | Autointerpreted label |
|---|---|---|
| 5 | 0 | conjunctions and relational words that suggest connections or similarities |
| 5 | 3 | instances of the verb "to be" in various forms |
| 5 | 4 | terms related to data management and database operations |
| 5 | 11 | programmatic structures and error handling related to Just-In-Time (JIT) compilation in code |
| 5 | 12 | mathematical expressions and notations |
| 4 | 150 | punctuation marks that signify dialogue or quoted speech |
| 4 | 151 | mathematical symbols and constructs used in equations |
| 4 | 163 | programming language syntax elements related to data definitions and structures |
| 4 | 164 | code-related terms and syntax that pertain to variable declarations and dimensionality in programming and mathematical contexts |
| 4 | 168 | biographical information about individuals, particularly their achievements and background |
| 3 | 577 | mentions of theorems and mathematical concepts |
| 3 | 608 | abstract representations of mathematical constructs or logical relations |
| 3 | 626 | references to large data files or memory usage issues in the context of programming |
| 3 | 642 | terms related to histopathological staining and tissue preparation for analysis |
| 3 | 647 | references to scientific publications and their formatting |
| 2 | 2052 | references to fundamental concepts and basic knowledge |
| 2 | 2074 | mathematical expressions related to specific variable representations in equations |
| 2 | 2142 | specific dates and references to issues or volumes in a publication |
| 2 | 2207 | references to private or confidential information |
| 2 | 2522 | references to specific viral genome components and their roles in replication processes |
| 1 | 8427 | instances of the word "the." |
| 1 | 8881 | metrics related to hospital stay and surgery outcomes |
| 1 | 9221 | descriptive words conveying positive emotions and vibrant imagery |
| 1 | 10121 | references to distinct stages in a process or study |
| 1 | 10290 | elements related to research proposals and their structure |

While it may well be the case that the features subject to the higher number of loss functions appear to be more general, we're still quite far away from clear evidence that such a method can help identify basic meanings that word meanings could consist in.

To summarise this section, it seems that SAE features are quite some way from looking like the entities that compose word meanings. In fact, there is a sense in which the method by which SAE features are constructed suggests not a form of decompositional view but a form of *atomism*. Atomism, defended by Fodor and many others, is the view that each meaningful unit possesses its meaning independently of all other terms, such that it can possess this meaning regardless of whether other terms possess any such meaning (Fodor & Lepore, 1992, p. 32). In encouraging an SAE to identify a set number of directions in the activation space, and then cementing that decision in the further step of providing LLM-generated labels for those features, we end up with a flat dictionary where little can be said about whether there are any relations between features and no particular guarantee that some basic level of meaning is captured by them.[14,15]

4. **A return to holism?**

So as far as the LLM picture goes, there is reason to be skeptical that SAE features should serve as the basic meanings as part of a decompositional picture. Does this save the holistic picture that Grindrod et al. (forthcoming) argue for? The key motivation for such a picture was that meaning appears to be represented using meaning proximity as its basic relation: that words will be proximal to the extent that they are similar in meaning. The existence of features challenged that insofar as it looked like an alternative explanation was available: that words will possess the meanings that they do by being encoded with particular features. But we have seen that while features do play the role of being co-activated in order to capture LLM activation patterns, and while they are often interpretable in terms of what content caused them to activate, they can not plausibly play the role of basic meaning parts that explain why a given word embedding is represented as having the meaning that it does. Rather than view features as sitting at some more basic level than word embeddings, they should instead be treated as meaningful units on par with word embeddings. The only difference between the two is that they are not explicitly represented in the space via some stored embedding. It is important to note in this regard that features, like word embeddings, are arranged according to meaning proximity as well (Templeton et al., 2024).

Features can arguably be captured as part of the holistic picture, then, as enjoying the same kind of holistic relations that word embeddings do. But as we saw in section 1, Grindrod et al. argued that this holism has to take a particular form, where really the meaning proximity relations that hold between words could be captured in an undirected network where words that are more similar in meaning have stronger connections in the network. Viewing the meaning relations in this way better captures the fact that different networks of different dimensionalities could capture the same meaning relations. We might think then that the only modification we have to make to this picture is to add implicit features to the network. So we would have a network consisting in words and implicit features that possess meaning similarity relations towards one another. The only fly in the ointment comes in the countability of the features. Features as generated through SAEs are just directions in the LLM's high-dimensional space. But suppose that using an SAE we identify two features relatively close to one another. The continuity of space would seem to guarantee a third feature that lies between the two. This would obviously give rise to an infinite number of such features. But with an infinite number of features, it then becomes hard to see how we could capture this as a network of nodes connected to one another.

---

[14] (Leask et al., 2025) argues that there is no canonical or privileged set of features that an SAE would ideally track. Through two studies, they show that smaller SAEs are incomplete, in that they lack features that larger SAEs possess and that would improve their performance if they did have them, and further that larger SAEs possess features that are non-atomic in that they can be decomposed into features possessed by smaller SAEs. Their argument is somewhat preliminary, focusing as it does on the behaviour of just two SAEs, so I won't rely on their conclusion here.

[15] (Smith, 2024) argues in a similar fashion that standard methods for generating SAE features ignores various possibilities regarding the relationship between SAEs, including that some SAE features may be composites of others.

But while the format that features and words are represented in is continuous, our intuitions about meaning suggest otherwise. To use our earlier examples from GEMMASCOPE_RES_16K, we shouldn't expect that lying exactly halfway between the feature for *salt* and the feature for *air conditioning systems* we will find a third meaningful feature that would somehow respect meaning proximity. Or, to take an example from the interactive map of features from (Templeton et al., 2024), the *Helen Keller* feature and the *Jane Goodall* feature lie close together, but we shouldn't expect some further meaningful feature to lie between the two. As Fodor worried about when it came to the residuum problem, we seem to be in the realm of impossible meanings here. Our intuitions about how meaning works suggest that the space, despite appearances should really be discrete, or at least partially so.[16] As long as that is the case, the possibility of understanding features and words as sat within a meaning proximity network is still on the table.

## 5. Conclusion

In this paper, I have considered the LLM picture of meaning in light of recent work on features generated from SAEs. I have outlined a way in which SAEs may be thought of as suggesting a feature composition view of meaning where word meanings are understood as composed of features (or feature meanings). I have argued that closer inspection of SAE features reveal that this feature composition is not currently plausible, and that Grindrod et al.'s holistic picture can accommodate the discovery of such features.

---

[16] Partially, because there are certain meaning phenomena that look continuous rather than discrete e.g. contextual thresholds for gradable adjectives such as "tall" is arguably a continuous affair.

*Transformer Circuits Thread.* https://transformer-circuits.pub/2024/scaling-monosemanticity/index.html

Vaswani, A., Shazeer, N., Parmar, N., Uszkoreit, J., Jones, L., Gomez, A. N., Kaiser, L., & Polosukhin, I. (2017). *Attention is all you need.* https://doi.org/10.48550/arxiv.1706.0376222